\documentclass{article}

\usepackage{amssymb}
\usepackage{arxiv}
\usepackage{float}
\usepackage{subfig}
\usepackage[utf8]{inputenc} 
\usepackage[T1]{fontenc}    
\usepackage{hyperref}       
\usepackage{url}            
\usepackage{booktabs}       
\usepackage{amsfonts}       
\usepackage{nicefrac}       
\usepackage{microtype}      
\usepackage{lipsum}
\usepackage{fancyhdr}       
\usepackage{graphicx}       
\graphicspath{{data/}}     
\usepackage{multirow}
\usepackage[normalem]{ulem}
\useunder{\uline}{\ul}{}
\usepackage{arydshln}

\title{ClinText-SP and RigoBERTa Clinical: a new set of open resources for Spanish Clinical NLP}

\author{
  Guillem García Subies \\
  Universidad Carlos III de Madrid \\
  Instituto de Ingeniería del Conocimiento \\
  \texttt{100500844@alumnos.uc3m.es, guillem.garcia@iic.uam.es} \\
   \And
  Álvaro Barbero Jiménez \\
  Instituto de Ingeniería del Conocimiento \\
  \texttt{alvaro.barbero@iic.uam.es} \\
   \And
  Paloma Martínez \\
  Computer Science and Engineering Department\\
  Universidad Carlos III de Madrid\\
  \texttt{pmf@inf.uc3m.es} \\
}

\begin{document}

\maketitle

\begin{abstract}
We present a novel contribution to Spanish clinical natural language processing by introducing the largest publicly available clinical corpus, \textbf{ClinText-SP}, along with a state-of-the-art clinical encoder language model, \textbf{RigoBERTa Clinical}. Our corpus was meticulously curated from diverse open sources, including clinical cases from medical journals and annotated corpora from shared tasks, providing a rich and diverse dataset that was previously difficult to access. RigoBERTa Clinical, developed through domain-adaptive pretraining on this comprehensive dataset, significantly outperforms existing models on multiple clinical NLP benchmarks. By publicly releasing both the dataset and the model, we aim to empower the research community with robust resources that can drive further advancements in clinical NLP and ultimately contribute to improved healthcare applications.
\end{abstract}

\section{Introduction}

The growing demand for advanced natural language processing (NLP) applications in the clinical domain has spurred significant research into domain-specific language models. However, a persistent challenge remains: the scarcity of large, high-quality, publicly available datasets in Spanish that can drive progress in clinical NLP. In this work, we address this gap by presenting ClinText-SP, the largest open corpus of Spanish clinical texts, and by introducing RigoBERTa Clinical, a state-of-the-art clinical encoder language model specifically adapted for the Spanish clinical domain.

The primary objective of this paper is to demonstrate that high-quality clinical data, when properly curated and combined with domain-adaptive pretraining, can substantially improve model performance on clinical NLP tasks. Our study is motivated by the need for accessible resources that enable both researchers and practitioners to develop and deploy effective clinical NLP solutions. Moreover, as generative Large Language Models (LLMs) continue to gain prominence, our work serves as an important benchmark, showcasing that specialized encoder-based models can still offer competitive performance with lower operational costs.

We begin by providing background information on clinical NLP and reviewing relevant previous work. Section 2 discusses the existing literature, highlighting both the advances in general language modeling and the specific challenges encountered in the clinical domain. In Section 3, we describe the development of ClinText-SP, detailing the sources, preprocessing methods, and overall composition of the dataset. Section 4 outlines the design and training procedure of RigoBERTa Clinical, which leverages the powerful RigoBERTa 2 \cite{rigoberta, survey} architecture through domain-adaptive pretraining on ClinText-SP. The experimental evaluation is presented in Section 5, where we benchmark our model against existing Spanish and multilingual models across several clinical tasks. Section 6 presents an ablation study that isolates the contributions of different data components, further validating the quality and impact of our dataset. Finally, Section 7 concludes the paper and outlines promising avenues for future work.

Our contributions are twofold: first, we release ClinText-SP, a valuable resource that we hope will accelerate research in Spanish clinical NLP; and second, we present RigoBERTa Clinical, which establishes a new performance baseline for clinical language understanding in Spanish. By making both resources publicly available, we aim to foster further advancements in this critical area of research.

\section{Previous Work}

Before diving into the advances in Spanish clinical narrative processing, we must highlight some non-Spanish models that have been pivotal in advancing medical NLP. Encoder models such as ClinicalBERT \cite{huang2020clinicalbertmodelingclinicalnotes} and BioBERT \cite{biobert} have demonstrated strong performance in understanding clinical language. On a larger scale, models such as Med-PaLM—leveraging up to 540 billion parameters— \cite{singhal2022largelanguagemodelsencode}, MedAlpaca \cite{medalpaca} and Clinical Camel \cite{clinicalcamel} have set new records on benchmarks in medical question answering, with recent evaluations like those from MedExpQA \cite{MedExpQA} further highlighting their effectiveness. As comprehensively reviewed in \cite{zhou2024surveylargelanguagemodels}, these developments illustrate that scaling model size significantly enhances the ability to capture and generate nuanced medical knowledge, thereby broadening the horizons of clinical applications in NLP.

Previous work in Spanish clinical narrative processing has focused on four main areas: (1) Building language models trained from scratch, (2) Fine-tuning pretrained LLMs using medical data, (3) Applying LLMs to information extraction tasks, and (4) Developing specific medical corpora used for training and testing systems.

One line of work has concentrated on creating language models specifically designed for clinical narratives. For instance, the BIO-EHR and EHR language models introduced in \cite{carrino-etal-2022-pretrained} were developed using a RoBERTa base model (with 12 self-attention layers as described in \cite{liu2019roberta}). These models were trained on a substantial dataset comprising 95M tokens from EHR sources—including discharge reports, clinical course notes, and X-ray reports—as well as 1.1B tokens from biomedical texts (scientific publications, EMEA documents, Wikipedia, patents, etc.). In the survey presented in \cite{survey} these models were compared against various general-domain encoder-based models (such as RigoBERTa, XLM-RoBERTa, and Galén models) along with generative models in NER tasks.

\cite{garcia2024medical} compiled the largest multilingual medical corpus—covering English, French, Italian, and Spanish—to train an open-source text-to-text model (mT5). The Spanish data sources include EMEA, SPACC, Wikimed, and PudMed, although no clinical narrative is included in the dataset.

Other research has focused on adapting pretrained language models to meet specific clinical tasks. For example, \cite{garcia2024gpt} fine-tuned both BERT and GPT models for Named Entity Recognition (NER) using a set of 550 oncology clinical notes.

Generative models have also been adapted to the clinical domain:
\begin{itemize}
    \item \cite{del2025comparative} explored instruction-based fine-tuning by evaluating various prompt strategies for NER. This work utilized overlapping Spanish corpora such as Distemist \cite{distemist}, SympTemist \cite{lima2023overview}, and MedProcNER \cite{lima2023overviewb}.
    \item In the realm of summarization, \cite{lopez2024evaluation} presented an approach to summarizing Spanish radiologist reports using both fine-tuned and zero-shot configurations with BART and T5 models on a private dataset containing 15,508 reports (specifically the findings sections).
\end{itemize}
It is important to note that these approaches make use of short Spanish corpora that contain only limited clinical narrative.

A significant focus of research has been the creation and annotation of specialized clinical corpora:
\begin{itemize}
    \item \textbf{European Clinical Case Corpus (E3C):} \cite{e3c} This corpus includes resources in Spanish, Basque, Italian, French, and English. The Spanish part is based on SPACCC \cite{intxaurrondo2018resources}, annotated with clinical entities and temporal information following the THYME annotation guidelines.
    \item \cite{survey} provides a detailed description of 17 Spanish corpora or datasets, including SPACCC, European Clinical Case Corpus (E3C), MEDDOCAN \cite{marimon2019automatic}, and the BARR corpus \cite{intxaurrondo2018resources}, among others.
    \item New clinical corpora are emerging, such as:
    \begin{itemize}
        \item \textbf{Nut Allergy Corpus} \cite{gonzalez2025clinical}, composed of 828 clinical notes annotated with semantic entities related to allergies.
        \item \textbf{Sec-CodiEsp Corpus} \cite{de2023open}, a subset of the CodiEsp corpus, containing 1038 Spanish clinical case notes from several medical specialties labeled with medical entities and sections (e.g., personal history, family history, explorations).
    \end{itemize}
\end{itemize}

Additional resources include corpora of clinical dialogues:
\begin{itemize}
    \item \textbf{VIR-PAT-QA} \cite{arana2024virtual} comprises 129 doctor--patient dialogues linked to corresponding clinical records, totaling 6290 question--answer pairs (translated from a French corpus).
    \item \textbf{CLARA-MeD Comparable Corpus} \cite{campillos2022building} consists of approximately 25,000 pairs of original and simplified texts, ranging from drug leaflets and summaries of product characteristics to systematic review abstracts (including cancer-related information summaries and clinical trial announcements).
    \item Other domain-specific efforts, such as \cite{lopez2021transformers}, have applied transformer architectures for automatic clinical coding in oncology clinical cases extracted from Galén Oncology Information systems.
\end{itemize}

Finally, it must be noted that in recent years, international shared tasks have been organized to advance the analysis of Spanish medical texts. Notable examples include CCodiEsp CLEF eHealth \cite{miranda2020overview}, Cantemist, \cite{cantemist}, GenoVarDis \cite{aguero2024overview} and Testlink \cite{altuna2023overview} among others.

These initiatives have played a pivotal role in benchmarking and promoting innovation in the field.

\section{Corpus}

We constructed the largest open Spanish clinical corpus to date, referred to as \textbf{ClinText-SP}, which is publicly available \footnote{\url{https://huggingface.co/datasets/IIC/ClinText-SP}}. This corpus focuses on assembling and processing clinical cases that, although publicly accessible, have largely been overlooked by the NLP research community. It also incorporates datasets from shared tasks and clinical knowledge extracted from Wikipedia.

The corpus contains $26$M tokens across $37,077$ samples, with an average of $700$ tokens per sample, providing a substantial length for each entry. The collected data is categorized into three primary sources: medical journals, annotated corpora, and miscellaneous sources.

\subsection{Medical Journals}

Clinical cases were extracted from various Spanish-language journals specializing in clinical case reports. Next, a description of each journal and their clinical cases will be presented:

\begin{itemize}
    \item \textit{Revista de Casos Clínicos en la Atención Primaria de la Sociedad Andaluza de Medicina Familiar y Comunitaria} \cite{jart}: This journal compiles clinical cases from primary care. A total of $519$ clinical cases were collected from journal editions published between $2017$ and $2023$.
    \item \textit{Atención Integral al Adolescente, Casos Clínicos} \cite{atadolcc}: This journal focuses on clinical cases involving adolescent patients. A total of $48$ clinical cases were collected from the $2011$, $2014$, and $2017$ editions. Notably, this journal is not published annually.
    \item \textit{Revista Oficial de la Sociedad Española de Medicina de la Adolescencia (SEMA)} \cite{adolescere}: This journal covers adolescent medicine, but not all articles are clinical cases, requiring a selection process. A total of $49$ clinical cases were gathered from editions published between $2013$ and $2024$.
    \item \textit{Revista Española de Casos Clínicos en Medicina Interna (RECCMI)} \cite{reccmi}: This journal compiles clinical cases in internal medicine. A total of $377$ clinical cases were collected from editions published between $2016$ and $2024$.
    \item \textit{Revista de Casos Clínicos en Salud Mental} \cite{ccsaludmental}: This journal specializes in clinical cases related to mental health. A total of $49$ clinical cases were collected from editions published between $2013$ and $2020$.
    \item \textit{Revista de Medicina Clínica (Rev Med Clin)} \cite{medicinaclinica}: As this journal publishes various types of medical articles, a selection process was applied to extract clinical cases. A total of $67$ clinical cases were collected from editions published between $2019$ and $2023$.
    \item \textit{Revista de Neurología} \cite{ccneurologia}: This journal focuses on neurology and includes various types of articles, necessitating a filtering process. A total of $103$ clinical cases were collected.
    \item \textit{Revista Española de Enfermedades Digestivas} \cite{reed}: This journal specializes in digestive diseases and features different types of articles, requiring a selection process. A total of $330$ clinical cases were collected.
    \item \textit{Revista Electrónica de Portales Medicos.com} \cite{portalesmedicos}: As a general medicine journal, a selection process was necessary to identify clinical cases. A total of $1199$ clinical cases were collected from editions published between $2011$ and $2024$.
\end{itemize}

\subsubsection{Clinical Cases Preprocessing}

First, all available PDFs and HTMLs were gathered from these journals, filtered to retain only clinical cases, and subsequently parsed and cleaned. As the cases were peer-reviewed, deduplication was deemed unnecessary.

The scraping and cleaning processes were challenging due to the diverse formats employed by the journals. For example, some journals provided all articles within a single PDF, others offered one PDF per article, while some did not offer PDFs at all. This necessitated customized crawling approaches for each journal. After obtaining the papers, an \textit{ad hoc} heuristic filter (specific regexes tailored to each journal) was applied to isolate clinical cases specific to each journal.

Text cleaning involved removing extraneous formatting, HTML artifacts, and irrelevant information such as author names and affiliations (noting that the clinical cases were anonymized by default). In some cases, journals maintained consistent formatting across articles and volumes, allowing efficient processing using regular expressions. For others, the task required leveraging the capabilities of Qwen2.5 \cite{qwen2.5}, an LLM used for assisted cleaning. While the LLM approach proved effective and reduced development time, it incurred higher processing time and computational costs compared to heuristic methods.

\subsection{Annotated Corpora}

The corpus also includes annotated datasets from various shared tasks and other sources:

\begin{itemize}
    \item \textit{European Clinical Case Corpus (E3C)} \cite{e3c}: E3C is a freely available multilingual corpus (English, French, Italian, Spanish, and Basque) of semantically annotated clinical narratives to allow for the linguistic analysis, benchmarking, and training of information extraction systems. Only the Spanish part was used, after deduplication with other sources, $504$ cases were left.
    \item \textit{SPACCC: Spanish Clinical Case Corpus} \cite{spaccc}: The SPACCC corpus was created after collecting $1000$ clinical cases from SciELO (Scientific Electronic Library Online), an electronic library that gathers electronic publications of complete full text articles from scientific journals of Latin America, South Africa and Spain.
    \item \textit{Chilean Waiting List Corpus} \cite{chilean}: The corpus contains referrals for several specialty consultations from the waiting list in Chilean public hospitals, they are not long text nor clinical cases, but they were chosen given their real life application. In total, $9000$ texts were gathered from this source.
    \item \textit{CANTEMIST (CANcer TExt Mining Shared Task – tumor named entity recognition)} \cite{cantemist}: This corpus comes from a shared task, the data features a collection of oncological clinical case reports. $1301$ cases were collected after deduplication.
    \item \textit{LivingNER: Named entity recognition, normalization \& classification of species, pathogens and food} \cite{livingner}: This corpus comes from a shared task with $15457$ clinical cases.
    \item \textit{CARES: A Corpus for classification of Spanish Radiological reports} \cite{cares}: A collection of radiological reports, although not being clinical cases, they are very useful clinical data. In total $3219$ cases were collected.
    \item \textit{MEDDOPLACE (MEDical DOcument PLAce-related Content Extraction)} \cite{meddoplace}: $1000$ clinical cases used in a Named Entity Recognition shared task about key places identification.
    \item \textit{Cardiology Clinical Case Reports (CardioCCC) } \cite{cardioccc}: $508$ cardiology clinical case reports used as part of the corpus in the MultiCardioNER shared task.
    \item \textit{Medical Documents Profession Recognition (MEDDOPROF)} \cite{meddoprof}: Shared task with clinical cases from over 20 different specialties, after the deduplication, a total of $321$ cases were used.
\end{itemize}

Given the frequent reuse of foundational corpora in shared tasks, deduplication was critical. To address this, a fuzzy deduplication strategy similar to the one used for GPT3 \cite{gpt3} was employed, using MinHash \cite{leskovec2014mining} as the hashing algorithm. Witht he help of Python Langdetect \cite{langdetect}, samples in languages other than Spanish were removed to ensure linguistic consistency.

\subsection{Other sources}

Additional medical knowledge was incorporated from Wikipedia \cite{enfermedades_wiki_marzo_2024} and other medical textbooks. Although this subset represents a small portion of the corpus, it plays a crucial role in balancing raw and structured knowledge, complementing the detailed clinical cases.

\subsection{Collected corpus volumetry}

As a summary of the collected corpora, Table \ref{tab:sources} presents information on the number of samples, total tokens and tokens per samples in the gathered sources. As we can see, the Journal Clinical Cases have much longer samples in general, which means that more complex language can be developed, potentially allowing more room for learning to a model trained from this dataset.

\begin{table}[H]
\centering
\begin{tabular}{l l l l l}
\toprule
\textbf{Sources} & \textbf{Samples} & \textbf{Tokens} & \textbf{Tokens/Sample} & \textbf{Type} \\
\midrule
CCAP                 & $519$   & $\sim0.708M$ & $1364$        & Journal Clinical Cases \\
AIACC                & $48$    & $\sim0.089M$ & $1859$        & Journal Clinical Cases \\
SEMA                 & $49$    & $\sim0.148M$ & $3020$        & Journal Clinical Cases \\
RECCMI               & $377$   & $\sim0.692M$ & $1834$        & Journal Clinical Cases \\
CCSM                 & $49$    & $\sim0.201M$ & $4102$        & Journal Clinical Cases \\
RevMedClin           & $67$    & $\sim0.111M$ & $1658$        & Journal Clinical Cases \\
Neurología           & $103$   & $\sim0.381M$ & $3698$        & Journal Clinical Cases \\
REED                 & $330$   & $\sim0.249M$ & $756$         & Journal Clinical Cases \\
Portales Medicos     & $1199$  & $\sim4.377M$ & $3650$        & Journal Clinical Cases \\
E3C                  & $504$   & $\sim0.336M$ & $666$         & Clinical Cases \\
SPACCC               & $1000$  & $\sim0.583M$ & $583$         & Journal Clinical Cases \\
Chilean Waiting List & $9000$  & $\sim0.468M$ & $52$          & Speciality Consultations \\
CANTEMIST            & $1301$  & $\sim1.586M$ & $1219$        & Shared Task Clinical Cases \\
LivingNER            & $15457$ & $\sim11.84M$ & $766$         & Shared Task Clinical Cases \\
CARES                & $3219$  & $\sim1.107M$ & $343$         & Radiological Reports \\
MEDDOPLACE           & $1000$  & $\sim1.051M$ & $1051$        & Shared Task Clinical Cases \\
CardioCCC            & $508$   & $\sim0.803M$ & $1582$        & Shared Task Clinical Cases \\
MEDDOPROF            & $321$   & $\sim0.307M$ & $957$         & Shared Task Clinical Cases \\
Wikipedia            & $945$   & $\sim0.572M$ & $605$         & Wikipedia \\
\midrule
\textbf{Average}     & $1894$  & $\sim1.348M$ & $712$         & - \\
\textbf{Total}       & $35996$ & $\sim25.62M$ & -             & - \\
\bottomrule
\end{tabular}
\caption{Sources, their sizes and type of documents. Sources are in the same order as presented in the paper, for instance CCAP = Revista de Casos Clínicos en la Atención Primaria, etc.}
\label{tab:sources}
\end{table}

By combining diverse sources, this corpus achieves an optimal balance between long, well-structured clinical cases and concise, schematic clinical knowledge, while avoiding overly specialized biomedical language and preserving accessibility to common language use.

\section{Model} \label{sec:model}

To evaluate the effectiveness of ClinText-SP and enhance the performance of language models for clinical Spanish, we developed \textbf{RigoBERTa Clinical} \footnote{\url{https://huggingface.co/IIC/RigoBERTa-Clinical}}.

\subsection{Domain Adaptation}

Given the limited availability of raw clinical cases in Spanish, training a model from scratch was not deemed effective. Prior research indicates that adapting general-purpose models through domain-specific pretraining often yields better results than training specialized models from scratch with small data \cite{survey}.

Therefore, RigoBERTa Clinical was initialized with the pretrained weights of RigoBERTa 2—a model with proven effectiveness in general Spanish and clinical NLP tasks \cite{rigoberta, survey}—and further trained using masked language modeling (MLM) \cite{devlin2019bert} on ClinText-SP. This approach leverages the linguistic knowledge encoded in RigoBERTa 2 while aligning the model to the clinical domain.

\subsection{Dataset Preprocessing}

We employed the tokenizer from RigoBERTa 2, ensuring consistency with the base model. To address the length of some clinical cases, which exceeded the maximum token length of $512$, we implemented a stride of $128$ tokens to segment long cases into overlapping samples. Shorter samples were padded as required. Out-of-vocabulary (OOV) words were managed using the same subword tokenization approach as RigoBERTa 2, ensuring the model's ability to generalize to unseen clinical terms.

\subsection{Training Procedure}

The pretraining procedure was conducted over two epochs. A grid search was performed to explore the interplay between batch size and learning rate, as smaller batch sizes typically require lower learning rates to stabilize training. The ranges for batch size and learning rate (Table \ref{tab:pretraining}) were chosen taking into account this fact and that learning rates should be smaller in a domain adaptation than in the first pre-training. The rest of the parameters were fixed to the default ones.

\begin{table}[H]
\centering
\subfloat[Parameter grid used for the training]{
\begin{tabular}{ll}
\hline
\textbf{Batch Size} & \textbf{Learning Rates} \\ \hline
$32$                & $\{5e-6, 1e-5, 2e-5\}$  \\
$64$                & $\{1e-5, 2e-5, 4e-5\}$  \\
$128$               & $\{1e-5, 4e-5, 8e-5\}$  \\ \hline
\end{tabular}}
\quad
\subfloat[Hyperparameters used for the training]{
\begin{tabular}{ll}
\hline
\textbf{Metaparameter} & \textbf{Configurations} \\ \hline
Warmup Steps           & $0$                     \\
Weight Decay           & $0.1$                   \\
Optimizer              & \textit{AdamW}          \\
Epochs                 & $2$                     \\
Adam Epsilon           & $1e-6$                  \\
Adam Beta2             & $0.98$                  \\ \hline
\end{tabular}}
\caption{Parameters used for the pre-training}
\label{tab:pretraining}
\end{table}

\subsection{Model Validation}

Given that pretraining loss does not reliably predict downstream performance \cite{liu2022pretraininglossbetterdownstream}, the best model was selected based on performance on downstream tasks. For this purpose, a subset of clinical datasets (livingner1, meddocan, and socialdisner) was used for validation, as these datasets are representative of the benchmark used in Section \ref{sec:experiments} and they are big enough to be statistically significant.

Fine-tuning was conducted on these datasets using a small grid search. A list of candidate models was chosen from the pretraining checkpoints starting from the first epoch and for every hyperparameter combination.

Then, candidate models were trained with this small grid search and ranked based on a weighted combination of metrics. The selected metrics were, average performance (weight of 0.8), standard deviation (weight of 1), median performance (weight of 1), best score obtained (weight of 0.75), worst score obtained (weight of 0.75), best 95\% of scores (weight of 1.1)  and 5\% worst scores (weight of 1.1). This comprehensive ranking ensured the selection of both the best-performing and most stable model across the range of the parameters in the grid search.

In the end, the best hyperparameters were: Batch size $32$, Learning rate $2e-5$ and $2800$ training steps which equal to $1.8$ epochs.

\subsection{Training Infrastructure}

Training was conducted using a single NVIDIA A100 GPU with 80GB of memory.

\section{Experiments} \label{sec:experiments}

RigoBERTa Clinical was evaluated against the latest Spanish and multilingual language models across a range of Spanish clinical datasets, adhering to the methodology outlined in \cite{survey}. As the same experimental framework was employed, the fine-tuning results of RigoBERTa Clinical are directly comparable to those reported in the aforementioned study. To facilitate reproducibility, the fine-tuning hyperparameters are detailed in Appendix \ref{appendix:meta}, and all fine-tuned versions of RigoBERTa Clinical are made publicly accessible in Appendix \ref{appendix:open}.

The benchmark utilized encompasses a diverse collection of publicly available annotated clinical corpora in Spanish. These datasets focus on two primary downstream tasks: multilabel classification and Named Entity Recognition (NER). Performance evaluation is conducted using the F1 score for NER datasets and the micro-averaged F1 score for classification tasks.

Table \ref{tab:results} presents the performance of RigoBERTa Clinical in comparison to the results reported in \cite{survey}. RigoBERTa Clinical achieves superior performance on the majority of datasets, consistently surpassing the previous state-of-the-art model, RigoBERTa 2. This improvement is reflected in an average performance increase of 0.01, which, given the consistency and size of the datasets, can be considered statistically significant.

\begin{table}[h!]
\centering
\makebox[\textwidth][c]{%
\begin{tabular}{l|ccccccc|cc}
\hline
                                 & \multicolumn{7}{c|}{\textit{Spanish Only models}}                                                               & \multicolumn{2}{c}{\multirow{2}{*}{\textit{Multilingual}}}  \\
                                 & \multicolumn{4}{c:}{\textit{Clinical models}}                           & \multicolumn{3}{c|}{\textit{General Models}} & \multicolumn{2}{c}{}        \\ \hline
Corpus                           & RigoCL               & XLM-RG  & BETO\_G & \multicolumn{1}{l:}{bsc-ehr} & BETO  & MarIA  & RigoB2               & mDeB3                &XLM-RL       \\ \hline
cantemist \cite{cantemist}       & {\ul \textbf{0.906}} & 0.898   & 0.802   & \multicolumn{1}{l:}{0.864}   & 0.898 & 0.902  & 0.903                & 0.890                & {\ul 0.904} \\
caresA \cite{cares}              & 0.991                & 0.989   & 0.977   & \multicolumn{1}{l:}{0.991}   & 0.992 & 0.992  & {\ul \textbf{0.997}} & 0.993                & {\ul 0.994} \\
caresC \cite{cares}              & {\ul \textbf{0.872}} & 0.823   & 0.551   & \multicolumn{1}{l:}{0.862}   & 0.835 & 0.840  & 0.854                & 0.756                & {\ul 0.847} \\
ctebmsp \cite{ctebmsp}           & {\ul \textbf{0.916}} & 0.881   & 0.726   & \multicolumn{1}{l:}{0.876}   & 0.880 & 0.877  & 0.907                & 0.902                & {\ul 0.906} \\
distemist \cite{distemist}       & {\ul \textbf{0.833}} & 0.759   & 0.346   & \multicolumn{1}{l:}{0.759}   & 0.801 & 0.793  & 0.832                & 0.808                & {\ul 0.817} \\
ehealth\_kd \cite{ehealthkd}     & {\ul \textbf{0.878}} & 0.830   & 0.658   & \multicolumn{1}{l:}{0.836}   & 0.843 & 0.836  & 0.865                & 0.844                & {\ul 0.871} \\
livingner1 \cite{livingner}      & {\ul \textbf{0.953}} & 0.907   & 0.646   & \multicolumn{1}{l:}{0.938}   & 0.938 & 0.939  & 0.951                & {\ul \textbf{0.953}} & 0.949       \\
livingner3 \cite{livingner}      & {\ul \textbf{0.698}} & 0.500   & 0.000   & \multicolumn{1}{l:}{0.604}   & 0.626 & 0.644  & 0.621                & 0.153                & {\ul 0.606} \\
meddocan \cite{meddocan}         & {\ul \textbf{0.982}} & 0.947   & 0.682   & \multicolumn{1}{l:}{0.967}   & 0.957 & 0.977  & 0.979                & 0.974                & {\ul 0.978} \\
pharmaconer \cite{pharmaconer}   & {\ul \textbf{0.931}} & 0.915   & 0.708   & \multicolumn{1}{l:}{0.904}   & 0.908 & 0.914  & 0.927                & 0.922                & {\ul 0.924} \\
socialdisner \cite{socialdisner} & 0.938                & 0.919   & 0.777   & \multicolumn{1}{l:}{0.921}   & 0.915 & 0.920  & {\ul \textbf{0.943}} & 0.935                & {\ul 0.941} \\ \hline
Average                          & {\ul \textbf{0.900}} & 0.852   & 0.625   & \multicolumn{1}{l:}{0.867}   & 0.872 & 0.876  & 0.889                & 0.830                & {\ul 0.885} \\
Wins group                       & {\ul \textbf{10}}    & 0       & 0       & \multicolumn{1}{l:}{0}       & 0     & 0      & 2                    & 1                    & {\ul 11}    \\
Wins total                       & {\ul \textbf{10}}    & 0       & 0       & \multicolumn{1}{l:}{0}       & 0     & 0      & 2                    & {\ul 1}              & 0           \\ \hline
\end{tabular}   
}
\caption{Results with models grouped into Spanish Only and Multilingual models. Spanish Only models are splitted into Clinical models and General models. Best results for every group are \uline{underlined} and best results overall are in \textbf{bold}. We also report the average of the scores for every model, wins in own group and wins overall. The metric is micro F1 for classification models and F1 for NER models. XLM-RG = XLM-R\_Galén \cite{galen}, BETO\_G = BETO\_Galén \cite{galen}, bsc-ehr = bsc-bio-ehr-es \cite{bsc}, BETO = BETO \cite{beto}, MarIA = MarIA \cite{maria}, RigoB2 = RigoBERTa 2 \cite{rigoberta}, mDeB3 = mDeBERTaV3 \cite{he2021debertav3}, XLM-RL = XLM-RoBERTa-Large \cite{xlmrcc100}, RigoCL = RigoBERTa Clinical}
\label{tab:results}
\end{table}

\begin{figure}[H]
\centering
     \includegraphics[width=\textwidth]{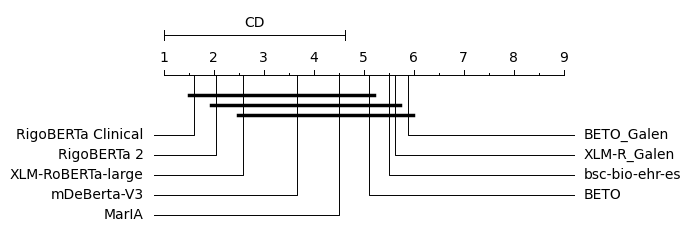}
      \caption{Nemenji plot of the evaluated models}
       \label{fig:nemenji}
\end{figure}

Figure \ref{fig:nemenji} provides a Nemenyi plot that ranks the models according to the benchmark results. According to this ranking, RigoBERTa Clinical emerges as the top performing model. It must be noted that RigoBERta Clinical ranks notably above the models trained only with clinical or medical data such as BETO\_Galen, XLM-R\_Galen and bsc-bio-ehr-es, which shows that our domain adaptation technique works better thanks to the general domain knowledge of the model.

However, it is important to note that, due to the limited number of datasets and the broad range of models assessed, the critical distance in the ranking remains substantial.

\section{Ablation Study}

This study aims to isolate and evaluate the contributions of different components of ClinText-SP to the overall performance of the proposed model. Specifically, we assess whether pretraining on only clinical cases from medical journals or only annotated corpora from shared tasks leads to meaningful improvements over the baseline model, RigoBERTa 2. By conducting this ablation study, we seek to determine whether each subset independently contributes to the final performance of RigoBERTa Clinical. In addition, our objective was to assess the overall quality and usefulness of the newly gathered data.

To this end, two new models were trained: one using only clinical cases extracted from medical journals, and the other using only annotated corpora from shared tasks. The training procedure remained consistent with the approach described in Section \ref{sec:model}, with the only difference being the use of ablated data subsets.

We hypothesize that both ablated models will outperform the baseline model, RigoBERTa 2, since they are domain-adapted versions. However, we also expect RigoBERTa Clinical to surpass both ablated models, as it benefits from exposure to a broader and more diverse dataset, allowing for better generalization.

Table \ref{tab:results-abl} presents a comparative evaluation of RigoBERTa 2, RigoBERTa Clinical, and the two ablated models in multiple benchmark data sets.

\begin{table}[H]
    \centering
    \renewcommand{\arraystretch}{1.2}
    \begin{tabular}{lcccc}
        \toprule
        \textbf{Dataset} & \textbf{RigoB2} & \textbf{RigoCL} & \textbf{RigoB+CC} & \textbf{RigoB+ST} \\
        \midrule
        socialdisner  & 0.943 & -0.005 & -0.008 & -0.006 \\
        meddocan      & 0.979 & +0.004 & +0.003 & +0.003 \\
        livingner1    & 0.951 & +0.003 & +0.002 & +0.004 \\
        \bottomrule
    \end{tabular}
    \caption{Performance comparison between the baseline model (RigoBERTa 2, RigoB2), the fully trained model (RigoBERTa Clinical, RigoCL), and the two ablated models trained exclusively on either clinical cases (RigoB+CC) or shared task corpora (RigoB+ST). F1 scores are shown for RigoBERTa 2, the rest of the models show the difference from the baseline.}
    \label{tab:results-abl}
\end{table}

Upon initial inspection, the results do not appear to fully align with expectations. Notably, RigoBERTa 2 achieves the highest score on the SocialDisNER dataset \cite{socialdisner}, outperforming both RigoBERTa Clinical and the ablated models (and also with the best result overall in the benchmark, see Table \ref{tab:results}). While seemingly counterintuitive, this outcome can be attributed to the nature of SocialDisNER, which consists primarily of tweets. Since RigoBERTa 2 was trained on general Spanish text, it retains a stronger representation of social media language, whereas RigoBERTa Clinical and the ablated models—having been further adapted to clinical literature—may have lost some of this general linguistic knowledge.

In contrast, for the Meddocan and LivingNER1 datasets, both ablated models outperform the baseline, confirming that the newly incorporated data positively impacts clinical domain adaptation. Furthermore, RigoBERTa Clinical achieves the best overall performance, demonstrating that the combination of both dataset components contributes to a more robust and generalizable clinical language model. It is also notable that the model trained on shared task corpora (RigoB+ST) achieves slightly better results than the one trained exclusively on clinical cases (RigoB+CC), likely due to the presence of SocialDisNER data in its pretraining set.

These findings suggest that both clinical cases from medical journals and annotated corpora from shared tasks play a significant role in enhancing the performance of RigoBERTa Clinical.

\section{Conclusion and Future Work}

In this work, we have introduced \textbf{ClinText-SP}, the largest publicly available Spanish clinical corpus, demonstrating that valuable clinical text data can be gathered from various open sources, even when not originally structured for NLP applications. By carefully curating and preprocessing this corpus, we have provided the community with a high-quality dataset that can facilitate further advancements in Spanish clinical NLP.  

Using this dataset, we trained \textbf{RigoBERTa Clinical}, the most effective Spanish clinical encoder-based language model to date. Our model significantly outperforms existing alternatives in multiple benchmark evaluations, demonstrating the value of domain-adaptive pretraining for clinical language understanding. To encourage further research, we have publicly released both the dataset and the trained model.  

Despite these advances, there is still considerable work to be done. The rapid progress of generative LLMs suggests that in the near future they may surpass fine-tuned encoder-based models in many clinical NLP tasks, especially with effective few-shot learning strategies. However, the high computational cost of these models remains a major limitation for real-world deployment in healthcare settings.  

Additionally, our work highlights the need for further architectural improvements, as we are reaching a point of diminishing returns. Recent advances, such as those proposed in ModernBERT \cite{modernbert}, suggest promising modifications that could enhance the efficiency and adaptability of encoder-based models for domain-specific applications. Future work should explore these architectural refinements, as well as more sophisticated domain adaptation techniques.  

Finally, we acknowledge that, while our dataset is the largest of its kind, there may still be biases in the data sources and gaps in domain coverage. Expanding the dataset to include more diverse clinical texts, as well as exploring multimodal approaches that integrate structured medical data, could further improve the robustness and applicability of Spanish clinical NLP models.  

We hope that our contributions, both in data and model development, serve as a foundation for further progress in Spanish clinical NLP, ultimately enabling more effective and accessible AI-driven tools for the healthcare community.  

\section*{Acknowledgments} 

This work was supported in part by the Instituto de Ingeniería del Conocimiento and Grant PID2023-148577OB-C21 (Human-Centered AI: User-Driven Adapted Language Models) by MICIU/AEI/ 10.13039/501100011033 and by FEDER/UE.

\bibliographystyle{unsrt}  
\bibliography{references}

\appendix

\section{fine-tuning Metaparameters}\label{appendix:meta}

RigoBERTa Clinical was fine-tuned using a grid search strategy with the grid shown in \ref{tab:metaparameters}

\begin{table}[H]
\centering
\begin{tabular}{ll}
\hline
\textbf{Metaparameter}  & \textbf{Configurations}    \\ \hline
Batch Size              & $16$                       \\
Classifier Dropout      & $\{0.08, 0.1\}$            \\
Learning Rate           & $\{2e-5, 5e-5\}$           \\
Warmup Steps            & $0$                        \\
Weight Decay            & $\{0, 1e-6, 1e-4\}$        \\
Optimizer               & \textit{AdamW Torch Fused} \\
Epochs                  & $10$                       \\
Early Stopping Patience & $3$                        \\ \hline
\end{tabular}
\caption{Parameter grid used for the experiments}
\label{tab:metaparameters}
\end{table}

\section{Public Resources}\label{appendix:open}

Every fine-tuned model follows the structure \texttt{{RigoBERTa-Clinical}-{corpus}}, so for example, the model id for the CANTEMIST corpus would be \texttt{IIC/RigoBERTa-Clinical-cantemist} \footnote{\url{https://huggingface.co/IIC/RigoBERTa-Clinical-cantemist}}. So, combining the model name and the corpus names from the list below leads to their corresponding Hugging Face Hub address.

The list below shows the availability of the fine-tuning corpora:

\begin{itemize}
    \item cantemist: \texttt{PlanTL-GOB-ES/cantemist-ner}
    \item caresA: \texttt{chizhikchi/CARES}
    \item caresC: \texttt{chizhikchi/CARES}
    \item ctebmsp: \texttt{lcampillos/ctebmsp}
    \item distemist: \texttt{bigbio/distemist}
    \item ehealth\_kd: \texttt{ehealth\_kd}
    \item livingner1: \texttt{IIC/livingner1}
    \item livingner3: \texttt{IIC/livingner3}
    \item meddocan: \texttt{bigbio/meddocan}
    \item pharmaconer: \texttt{PlanTL-GOB-ES/pharmaconer}
    \item socialdisner: \texttt{IIC/socialdisner}
\end{itemize}

\end{document}